\documentclass{article} % For LaTeX2e
\usepackage{iclr2023_conference_tinypaper,times}

\usepackage{amsmath,amsfonts,bm}

\def\eqref#1{equation~\ref{#1}}
\def\1{\bm{1}}

\def\vx{{\bm{x}}}

\DeclareMathAlphabet{\mathsfit}{\encodingdefault}{\sfdefault}{m}{sl}
\SetMathAlphabet{\mathsfit}{bold}{\encodingdefault}{\sfdefault}{bx}{n}

\usepackage{makecell} % Xhline
\usepackage{multirow} % Table 기능
\usepackage{adjustbox} % \resizebox{\textwidth}{!}
\usepackage{hyperref}
\usepackage{url}
\usepackage{algorithm}
\usepackage{algorithmic}

\title{Colorful Cutout: Enhancing Image Data \\ Augmentation with Curriculum Learning}

\author{Juhwan Choi and Youngbin Kim \\
Chung-Ang University, Seoul, Republic of Korea \\
\texttt{\{gold5230, ybkim85\}@cau.ac.kr} \\
}

\iclrfinalcopy % Uncomment for camera-ready version, but NOT for submission.
\begin{document}
\maketitle

\begin{abstract}
Data augmentation is one of the regularization strategies for the training of deep learning models, which enhances generalizability and prevents overfitting, leading to performance improvement. Although researchers have proposed various data augmentation techniques, they often lack consideration for the difficulty of augmented data. Recently, another line of research suggests incorporating the concept of curriculum learning with data augmentation in the field of natural language processing. In this study, we adopt curriculum data augmentation for image data augmentation and propose \textbf{colorful cutout}, which gradually increases the noise and difficulty introduced in the augmented image. Our experimental results highlight the possibility of curriculum data augmentation for image data. We publicly released our source code to improve the reproducibility of our study. % \footnote{\url{https://anonymous.4open.science/r/COLORFUL_CUTOUT_AUG}}
\end{abstract}

\section{Introduction}

Data augmentation is an important regularization trick to train the deep learning model that aims to improve generalization ability and prevent overfitting \citep{yang2022image}. From the basic manipulation of input images, such as cropping, rotating, and jittering, data augmentation techniques for image data have evolved. For example, cutout and random erasing \citep{devries2017cutout, zhong2020random} augmentation suggested a dropout strategy on the input image level by erasing a portion of a given image. After mixup \citep{zhang2018mixup} introduced the concept of vicinal risk minimization through the mixture of two images, cutmix \citep{yun2019cutmix} proposed a strategy that combines cutout and mixup.

However, previous approaches have limited considerations about the difficulty of augmented data. It is widely accepted that a well-defined training procedure with the consideration of the difficulty of given data can enhance the performance of the trained model \citep{bengio2009curriculum, soviany2022curriculum}. Recently, researchers have been exploring the combination of data augmentation and curriculum learning in the context of curriculum data augmentation \citep{wei2021few, ye2021efficient, lu2023pcc}. Nonetheless, these approaches are mainly performed for the text data.

In this paper, we propose a novel curriculum data augmentation technique for image data. Specifically, it first introduces the colorization into cutout, which originally erases the portion of a given image. Additionally, through the division of the erasure box and filling the sub-regions with different colors, we are allowed to adjust the difficulty of the augmented image. To the best of our knowledge, this is the first study that pioneers curriculum data augmentation in the computer vision field. Our comprehensive experiment on various models and datasets demonstrates the effectiveness of our method, highlighting the advantage of curriculum data augmentation.

\begin{figure*}[h]
    \centering
    \includegraphics[width=0.95\textwidth]{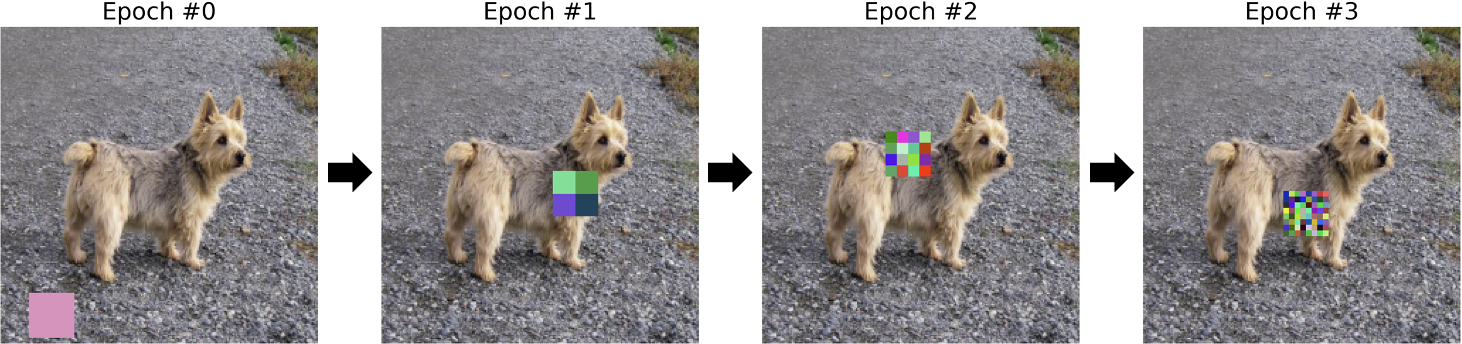}
    \caption{As the training procedure progresses, colorful cutout introduces more complex and difficult noise into augmented images.}
\label{fig-method}
\end{figure*}

\section{Method}

We first briefly explain the procedure of traditional cutout. From the given image $\vx$, it randomly selects a box region with a fixed size. After the selection, the box is erased and filled with zero value. Instead of simple erasure, our proposed colorful cutout fills the given box with a random color. This colorization establishes additional variation in augmented images, addressing a common limitation of previous methods and contributing to performance gain \citep{zhang2022good}.

Additionally, colorful cutout introduces the concept of curriculum data augmentation through the division of the erasure box into sub-regions. Each of these sub-regions could have different colors. As the number of sub-regions increases, the erasure box becomes more tangled, resulting in more difficult samples as the training progresses. Figure~\ref{fig-method} demonstrates the gradual increment of difficulty as the training progresses. Please refer to Appendix~\ref{app-algo} for the pseudo-code of colorful cutout.

\section{Experiment}

\begin{table}[t]
\caption{Accuracy (\%) for each model and augmentation techniques across three datasets. ``C10'', ``C100'', and ``TI'' symbolizes CIFAR-10, CIFAR-100, and Tiny ImageNet, respectively.}
\label{tab-main}
\begin{center}
\begin{tabular}{l|ccc|ccc|ccc}
\Xhline{3\arrayrulewidth}
                                                         & \multicolumn{3}{c|}{\textbf{ResNet50}} & \multicolumn{3}{c|}{\textbf{EfficientNet-B0}} & \multicolumn{3}{c}{\textbf{ViT-B/16}} \\
Dataset                                                  & C10         & C100         & TI        & C10           & C100           & TI           & C10         & C100        & TI        \\ \Xhline{2\arrayrulewidth}
Baseline                                                 & 94.82       & 80.56        & 73.09     & 96.48         & 82.38          & 78.25        & 95.58       & 83.94       & 81.54     \\
Cutout                                                   & 95.49       & 80.97        & 73.52     & 96.56         & 82.53          & 78.41        & 96.08       & 84.21       & 81.49     \\
Mixup                                                    & 95.56       & 81.15        & 73.24     & 96.63         & 82.50          & 78.26        & 96.45       & 84.25       & 81.48     \\
CutMix                                                   & 95.67       & 81.45        & 73.63     & 96.67         & 82.96          & 78.53        & 96.27       & 84.32       & 81.82     \\
\begin{tabular}[c]{@{}l@{}}Ours\\ w/o Curr.\end{tabular} & 95.16       & 81.15        & 73.61     & 96.72         & 82.92          & 78.32        & 96.35       & 84.20       & 82.15     \\
Ours                                                     &\textbf{95.70}&\textbf{81.57}&\textbf{73.81}&\textbf{96.81}&\textbf{83.37}&\textbf{78.65}&\textbf{96.55}&\textbf{84.36}&\textbf{82.36}         \\ \Xhline{3\arrayrulewidth}
\end{tabular}
\end{center}
\end{table}

We conducted an experiment to evaluate the effectiveness of our proposed method. First, we adopted three different datasets, CIFAR-10, CIFAR-100 \citep{krizhevsky2009learning}, and Tiny ImageNet \citep{le2015tiny} for evaluation. Second, we compared our methods against various previous augmentation techniques, including traditional cutout \citep{devries2017cutout}, mixup \citep{zhang2018mixup}, and cutmix \citep{yun2019cutmix}. Last, we applied these methods on three different models, CNN-based ResNet50 \citep{he2016deep} and EfficientNet-B0 \citep{tan2019efficientnet}, and Transformer-based ViT-B/16 \citep{dosovitskiy2020image}. Please refer to Appendix~\ref{app-implementation-details} for more details.

Table~\ref{tab-main} displays the experimental result. The results demonstrate a significant improvement in model performance with colorful cutout compared to other methods, particularly traditional cutout. Additionally, our ablation experiment on colorful cutout without the curriculum data augmentation shows similar performance to cutout, which suggests the curriculum data augmentation plays an important role for enhancing the performance of the model. This shows the potentiality of curriculum data augmentation in image data augmentation.

\section{Conclusion}

In this paper, we proposed a simple yet effective augmentation strategy that incorporates the concept of curriculum data augmentation into the computer vision field. The experimental results highlight the effectiveness of our approach and the possibility of curriculum image augmentation. Future research could investigate applying curriculum data augmentation to other image augmentation strategies and introducing soft labels to augmented data considering its difficulty \citep{choi2023softeda}.

\subsubsection*{Acknowledgements}
This research was supported by Basic Science Research Program through the National Research Foundation of Korea(NRF) funded by the Ministry of Education(NRF-2022R1C1C1008534), and Institute for Information \& communications Technology Planning \& Evaluation (IITP) through the Korea government (MSIT) under Grant No. 2021-0-01341 (Artificial Intelligence Graduate School Program, Chung-Ang University).

\subsubsection*{URM Statement}
First author Juhwan Choi meets the URM criteria of ICLR 2024 Tiny Papers Track. He is outside the range of 30-50 years, non-white researcher.

\bibliography{iclr2023_conference_tinypaper}
\bibliographystyle{iclr2023_conference_tinypaper}

\appendix

\section{Implementation Details}
\label{app-implementation-details}
This section describes implementation details and setups for reproduction. Please refer to our source code for more detailed information. \footnote{\url{https://github.com/c-juhwan/colorful-cutout-aug}}

\textbf{Model Implementation.} Every three models were based on the pre-trained checkpoints on ImageNet offered by TorchVision \citep{torchvision2016} library. After the feature extraction from the pre-trained models, a two-layer classification with a dropout layer of $p=0.2$ and ReLU activation is followed. 

Every input image is resized to 256$\times$256 and randomly cropped into 224$\times$224 size in the training procedure. In the validation and test procedure, an image with 224$\times$224 size is obtained from the center of the original image.

\textbf{Augmentation Implementation.} For traditional cutout, cutmix, and our method, the size of the box $w$ is set to 32$\times$32 for every model. For the mixup and cutmix method, we used $\alpha=0.2$, where $\alpha$ denotes the bound of the beta distribution that determines the mixup ratio.

Colorful cutout increases the number of sub-regions as the training epoch increases. Specifically, the number of sub-regions is defined as $2^{N_{\textit{epoch}}}$. Each sub-region is assigned different random colors. We set the initial $N_{\textit{epoch}}$ starts at 0, indicating that there is no sub-region in the first epoch. Please refer to Figure~\ref{fig-method} as an example.

\textbf{Datasets.} CIFAR-10 \cite{krizhevsky2009learning} is an image classification dataset consisting of 50,000 training images and 10,000 test images in 10 classes. CIFAR-100 is an extended version of CIFAR-10, which is composed of 100 classes. Tiny ImageNet \citep{le2015tiny} is a subset of ImageNet \citep{deng2009imagenet}, which has 200 classes and 100,000 training images. Three datasets were downloaded from Datasets library \citep{lhoest2021datasets} operated by Hugging Face. As there is no predefined validation set exists, we randomly selected 10\% of the training data as the validation set.

\textbf{Hyperparameters.} Adam \citep{kingma2015adam} has been deployed as the optimizer, with a learning rate of 5e-5. We trained each model for 5 epochs with a batch size of 32. We applied label smoothing \citep{szegedy2016rethinking} with a smoothing factor 0.05 for every model.

\textbf{Further Details.} Every experiment was performed using a single NVIDIA RTX 3090 GPU. We trained the model with our method for 75.7 minutes on Tiny ImageNet, while cutout baseline took 74.2 minutes.

\section{Comparison between Other Techniques}
We provide an example of colorful cutout compared to other methods on the same image.

\begin{figure*}[h]
    \centering
    \includegraphics[width=1\textwidth]{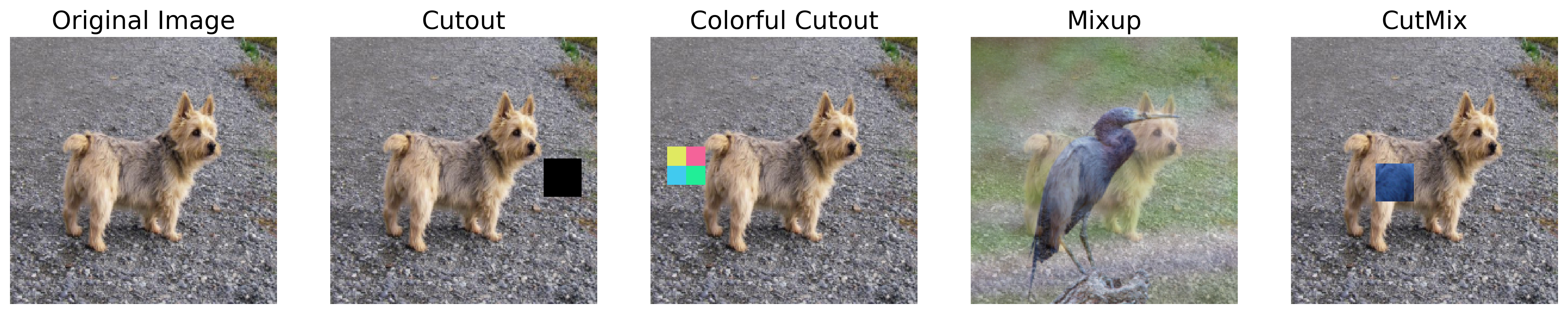}
    \caption{An example of our proposed colorful cutout compared to previous data augmentation methods.}
\label{fig-comparison}
\end{figure*}

\section{Algorithm of Colorful Cutout}
\label{app-algo}

We provide a pseudo-code for colorful cutout.

\begin{algorithm}
\caption{The procedure of colorful cutout.}
\begin{algorithmic}[1]
\REQUIRE Given image $\vx$, pre-defined size of erasure box $w$, current epoch index $N_{epoch}$
\STATE Randomly generate erasure box $B$ with size of $w \times w$ from $\vx$
\STATE Get the number of sub-region $N_{region} = 2^{N_{epoch}}$
\STATE Divide $B$ into $N_{region}$ squared sub-regions
\STATE Fill divided $B$ with $N_{region}$ random colors
\STATE Return augmented image $\hat{\vx}$
\end{algorithmic}
\end{algorithm}

\end{document}